%% file: acl_latex.tex
\pdfoutput=1

\documentclass[11pt]{article}

\usepackage{graphicx}
\usepackage{subfig}
\usepackage[review]{acl}

\usepackage{times}
\usepackage{latexsym}

\usepackage[T1]{fontenc}

\usepackage[utf8]{inputenc}

\usepackage{microtype}
\usepackage{soul}
\usepackage{tikz}
\def\checkmark{\tikz\fill[scale=0.4](0,.35) -- (.25,0) -- (1,.7) -- (.25,.15) -- cycle;} 
\definecolor{thistle}{RGB}{190,151,190}

\definecolor{coral}{RGB}{254,125,93}

\newcommand{\ours}[0]{\texttt{SPATIAL}}

\title{Pretraining on Interactions for Learning Grounded Affordance Representations}

\author{First Author \\
  Affiliation / Address line 1 \\
  Affiliation / Address line 2 \\
  Affiliation / Address line 3 \\
  \texttt{email@domain} \\\And
  Second Author \\
  Affiliation / Address line 1 \\
  Affiliation / Address line 2 \\
  Affiliation / Address line 3 \\
  \texttt{email@domain} \\}

\begin{document}
\maketitle
\begin{abstract}
Lexical semantics and cognitive science point to \textit{affordances} (i.e. the actions that objects support) as critical for understanding and representing nouns and verbs. However, study of these semantic features has not yet been integrated with the ``foundation'' models that currently dominate language representation research. We hypothesize that predictive modeling of object state over time will result in representations that encode object affordance information ``for free''. We train a neural network to predict objects' trajectories in a simulated interaction and show that our network's latent representations differentiate between both observed and unobserved affordances. We find that models trained using 3D simulations from our \ours\ dataset outperform conventional 2D computer vision models trained on a similar task, and, on initial inspection, that differences between concepts correspond to expected features (e.g., \textit{roll} entails \textit{rotation}). Our results suggest a way in which modern deep learning approaches to grounded language learning can be integrated with traditional formal semantic notions of lexical representations.
\end{abstract}

\input{sections/intro}

\input{sections/dataset}

\input{sections/model}

\input{sections/experiments}
\input{sections/related}
\input{sections/conclusion}
\section*{Acknowledgments}
This research is supported in part by DARPA via the GAILA program (HR00111990064). The views and conclusions contained herein are those of the authors and should not be interpreted as necessarily representing the official policies, either expressed or implied, of DARPA or the U.S.\ Government. Thanks to Chen Sun, Roman Feiman, members of the Language Understanding and Representation (LUNAR) Lab and AI Lab at Brown, and the reviewers for their help and feedback on this work.

\bibliography{acl_latex}
\bibliographystyle{acl_natbib}
\clearpage
\appendix

\section{Appendix A}
\label{sec:appendix}

\begin{figure}[ht!]
    \centering
    \includegraphics[ width=7.5cm ]{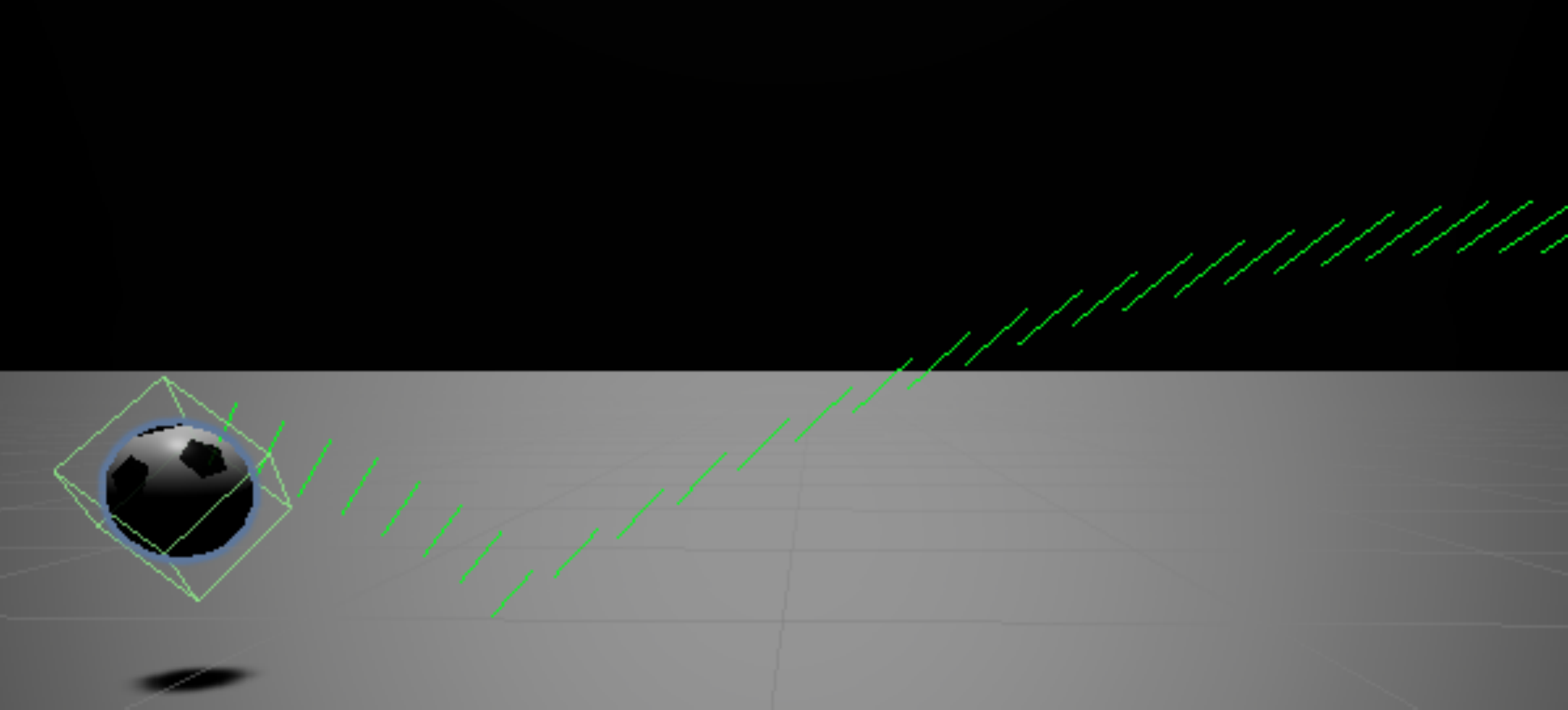}
    \caption{Example of an interaction from \ours. The model predicts the position of the soccer ball at future timesteps. To do that, it must encode some knowledge that soccer balls \texttt{bounce} and \texttt{roll}. As input, our model takes 9 3D points: the eight corners of the box surrounding the ball, plus the center point.}
    \label{fig:ex_sliding_obj}
\end{figure}

\begin{table*}[ht]
    \centering
    \begin{tabular}{c|c|c|c|c|c|c|c}
        \textbf{Object} & \textbf{Test set?} & \textbf{Slide} & \textbf{Roll} & \textbf{Stack} & \textbf{Contain} & \textbf{W-Grasp} & \textbf{Bounce} \\
        \hline
BombBall & \checkmark & & \checkmark & & & & \checkmark \\
EyeBall & & & \checkmark & & & & \checkmark \\
SpikeBall & & & \checkmark & & & & \\
Vase\_Amphora &  & & \checkmark & & & & \\
Vase\_Hydria & & & \checkmark & & & & \\
Vase\_VoluteKrater & \checkmark & & \checkmark & & \checkmark & & \\
book\_0001a & & \checkmark & & \checkmark & & & \\
book\_0001b & & \checkmark & & \checkmark & & & \\
book\_0001c & \checkmark & \checkmark & & \checkmark & & & \\
bowl01 & & \checkmark & \checkmark & \checkmark & \checkmark & & \\
cardboardBox\_01 & & \checkmark & & \checkmark & & & \\
cardboardBox\_02 & & \checkmark & & \checkmark & \checkmark & & \\
cardboardBox\_03 & \checkmark  & \checkmark & & \checkmark & & & \\
Cola Can & & \checkmark & \checkmark & \checkmark & & \checkmark & \\
Pen black & \checkmark  & & \checkmark & & & \checkmark & \\
Gas Bottle & \checkmark  & & \checkmark & & & & \\
Soccer Ball & & & \checkmark & & & & \checkmark \\
can small & & \checkmark & \checkmark & \checkmark & & \checkmark & \\
can & \checkmark & \checkmark & \checkmark & \checkmark & & \checkmark & \\
meat can box & & \checkmark & & \checkmark & & & \\
spam can & & \checkmark & & \checkmark & & \checkmark & \\
AtomBall & & & \checkmark & & & & \checkmark \\
Bottle2 & \checkmark & & \checkmark & & & \checkmark & \\
plate02 & \checkmark & \checkmark & & \checkmark & & & \\
plate02\_flat & & \checkmark & & \checkmark & & & \\
Bottle1 & & & \checkmark & & & \checkmark & \\
WheelBall & & & \checkmark & & & & \checkmark \\
wine bottle 04 & \checkmark & & \checkmark & & \checkmark & \checkmark & \\
coin & \checkmark & \checkmark & & \checkmark & & & \\
BuckyBall & \checkmark  & & \checkmark & & & & \checkmark \\
SplitMetalBall & \checkmark  & & \checkmark & & & & \checkmark \\
bowl02 & & \checkmark & \checkmark & \checkmark & \checkmark & & \\
bowl03 & \checkmark  & \checkmark & \checkmark & \checkmark & \checkmark & & \\
mug02 & & \checkmark & & & \checkmark & \checkmark & \\
mug03 & \checkmark  & \checkmark & & & \checkmark & \checkmark & \\
Old\_USSR\_Lamp\_01 & \checkmark & \checkmark & & & & \checkmark & \\
lamp & \checkmark  & \checkmark & \checkmark & & & \checkmark & \\
Ladle & & \checkmark & & & & \checkmark & \\
Apple & & & \checkmark & & & & \\
    \end{tabular}
    \caption{All objects in the dataset and their associated affordances}
    \label{tab:checkmark_table}
\end{table*}

\begin{table*}[t]
    \centering
    \begin{tabular}{c|c|c|c||c|c||c|c||c|c}
        \multicolumn{10}{c}{Binary Classification Accuracy of Affordance Probes (Random=50\%)}\\ \hline
        & N examples & \multicolumn{2}{c}{3D Blind} |& \multicolumn{2}{c}{3D Visual-Spatial} |& \multicolumn{2}{c}{2D Visual} |& \multicolumn{2}{c}{No-Training}\\ \hline
         Affordance &  & Acc. (\%) & F1 & Acc. (\%) & F1 & Acc. (\%) & F1 & Acc. (\%) & F1 \\ \hline
         
         Slide & 1715 & 59.2 & 62    & \textbf{63.7} & 67 & 59.6 & 62 & 61.2 & 64 \\
         Roll & 1258 & 66.5 & 66     & \textbf{69.0} & 70 & 56.0 & 58 & 66.3 & 66 \\
         Stack & 1307 & 64.7 & 65    & \textbf{65.7} & 63 & 58.0 & 58 & 63.4 & 63 \\
         Contain & 1510 & \textbf{66.5} & 68  & 65.2 & 67 & 58.2 & 63 & 59.3 & 64 \\
         W-Grasp & 1652 & \textbf{67.2} & 68  & 66.3 & 0.68 & 62.9 & 61 & 63.4 & 64 \\
         Bounce & 276 & \textbf{82.6} & 83  & 79.7 & 79 & 73.9 & 76.  & 76.0 & 75 \\
    \end{tabular}
    \caption{Results from probing experiments on RPIN compared to the unity models trained on the same amount of data. Because data was limited, we partition the data so that there is an even number of positive and negative examples in the test set for each affordance. Interaction based pretraining outperforms visual dynamics in all categories} 
    \label{tab:probe_tab}
\end{table*}


\begin{table*}[ht]
    \centering
    \begin{tabular}{c|c|c}
        Affordance & Number of Objects & Percentage of objects (\%)\\ \hline
        Slide & 22 & 56.41 \\
        Roll & 23 & 58.97 \\
        Stack & 17 & 43.59\\
        Contain & 8  & 20.51\\
        Wrap-grasp & 13 & 33.33 \\
        Bounce & 7 & 17.95
    \end{tabular}
    \caption{Each affordance we are interested in learning and the number and percentage of objects out of the 39 have a positive label for that affordance.} 
    \label{tab:aff_counts}
\end{table*}

\begin{figure*}[ht]
    \centering
    \includegraphics[scale=.08]{figures/unity_aff_all_objs_pdf.pdf}
    \caption{All objects that were used in training and testing. Some objects in the test set are visually similar to their training analogues, but differ in size and mass.}
    \label{fig:all_objs}
\end{figure*}

\subsection{Spatial Model Training Details}
For both of the 3D spatial data models, we train with an encoder-decoder transformer with one encoder and one decoder layer with one attention head. We found that changing the number of attention heads did not affect performance noticeably in either direction. We use a batch size of 64 and a transformer embedding dimension of 100. We use a feed-forward dimension of 200. We initialize with a learning rate of 1e-4. The models were trained to predict the next $k=8-16$ frames and we did not see a large benefit in training to predict longer sequences. We trained the models for 400 epochs although we notcied the ablated 3D Blind model tended to converge at or before 100 epochs across our experiments.

The beginning of the sequence, which was up to four seconds minus the $k$ prediction frames, was fed into the transformer encoder which encoded representations of dimension $e$. We averaged these output embeddings as input in our probing experiments. The $e$ embeddings were fed into the decoder network, which then predicts the next $k$ frames. We believe that training with longer sequences would be more beneficial for training a decoder-only model, which we would like to explore in future work. In preliminary experiments, we tested whether masking a proportion of frames in the encoder would be beneficial for the representation learning task. We saw a slight decrease in performance, and so did not perform a thorough analysis on the effect of masking.

\subsection{t-SNE Configuration}
We report a t-SNE of representations derived from our 3D Blind model and the 2D Visual model. The parameters for creating each t-SNE was similar but varied in a few ways:
\textbf{Common Hyperparameters:} learning rate: 200, iterations: 1000, stopping threshold of gradient norm: 1e-7
\textbf{3D Blind t-SNE specifics:} perplexity: 30, initialized randomly
\textbf{2D Visual specifics:} perplexity: 5, initialized with PCA. We found that random initialization was inconsistent in that it would sometimes cause small clouds of dense points to appear as their own clusters.

\section{Appendix B}
\label{sec:rpin_appendix}
\subsection{RPIN Training details}
We use a learning rate of 1e-3 with a batch size of 50 and train for a maximum of 20M iterations with 40,000 warmup iterations. Training data is augmented with random horizontal flips. Unlike in \citet{rpin} we don't use vertical flips because our videos contain objects falling due to gravity. One important difference is that at training time the model predicts 10 frames in the future, and at test time predicts 20 (as opposed to 20 and 40 respectively in Simulated Billiards). Within one video, our interactions seem more complex than one sequence in the Simulated Billiards dataset, so we introduced this difference to create more training examples.

\begin{table*}[]
    \centering
    \begin{tabular}{c|c}
    \multicolumn{2}{c}{\large  RPIN Model validation loss in the $t \in [T_{train}, 2 \times T_{train}]$ setting}\\\hline
    \multicolumn{1}{c}{Model} & \multicolumn{1}{c}{Loss (MSE)} \\ \hline
    SimB \citep{rpin} & 25.77 \\
    SimB (our results) & 15.53 \\
    Unity Videos & 20.98
    \end{tabular}
    \caption{Each affordance we are interested in learning and the number and percentage of objects out of the 39 have a positive label for that affordance.} 
    \label{tab:rpin_loss}
\end{table*}

\begin{figure*}[ht]
    \centering
    \includegraphics{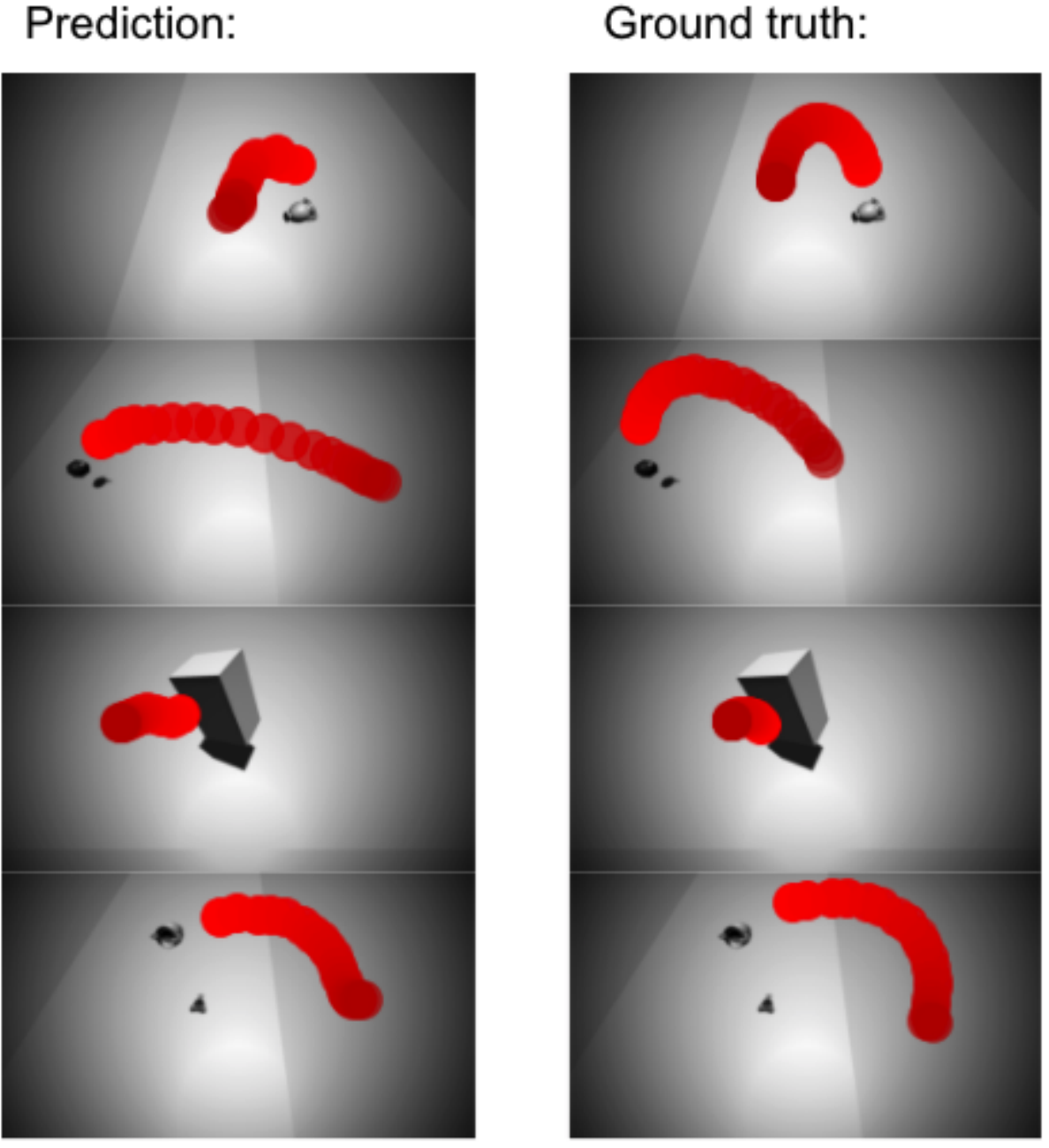}
    \caption{Results from the training of the RPIN visual dynamics model on videos of our Unity dataset interactions. Red circles show the predictions of the following center points of the bounding boxes of the object given the start of the interaction}
    \label{fig:rpin_sbs}
\end{figure*}

\section{Appendix C}
\label{sec:demo_appendix}

\subsection{Counterfactual Perturbations Setup}
\begin{figure*}[ht]
    \centering
    \includegraphics[width=\textwidth]{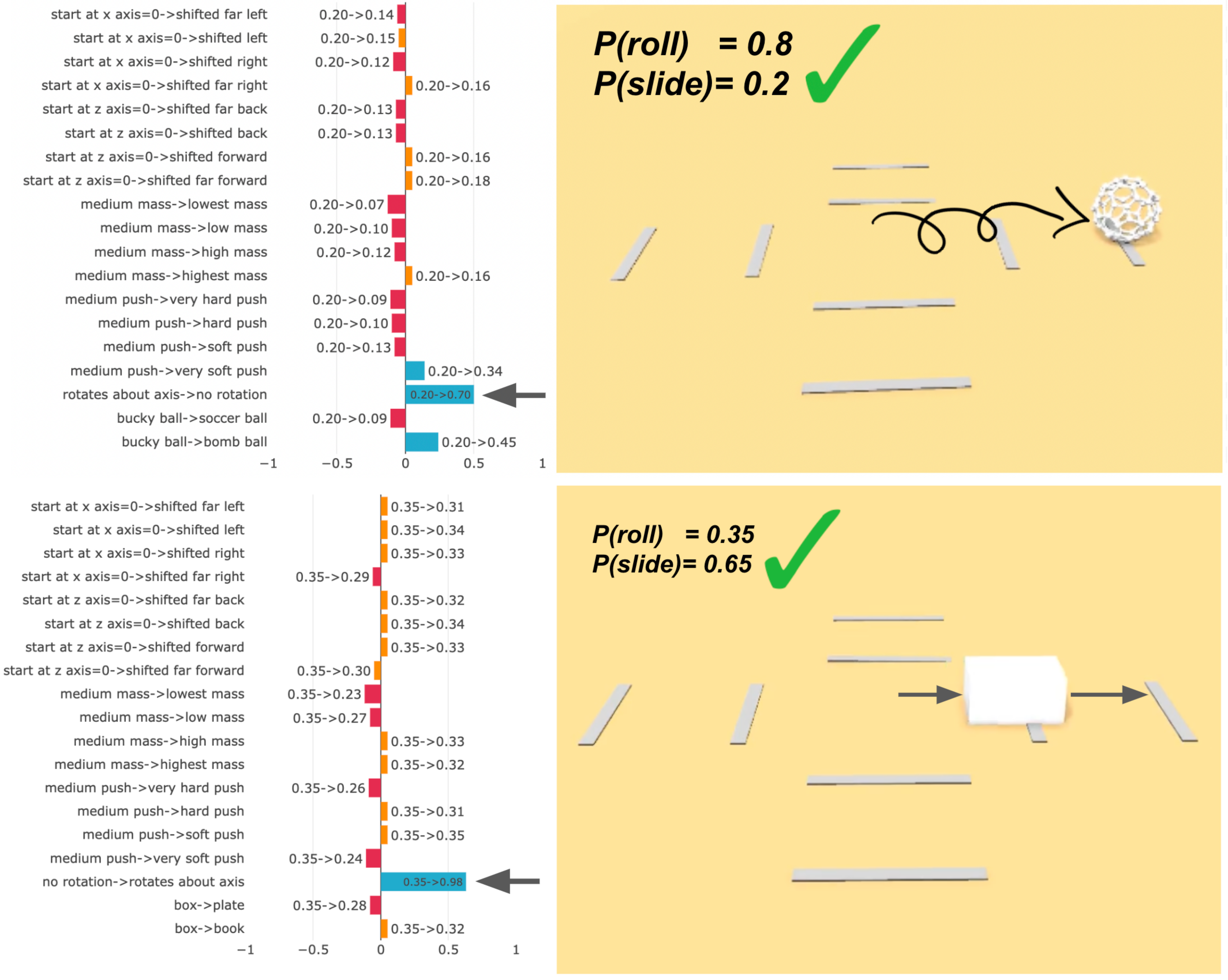}
    \caption{Two examples from the counterfactual analysis that show robustness to changing spurious features. The table in the top example displays the changes in probability in predicting the object as sliding. Conversely, the bottom example table shows the change in probability of predicting the object as rolling. Arrows in the left table indicate where the perturbation \textit{does} affect the label of the action (either by making the object able or unable to roll). In both cases, the probe correctly flips its prediction on the encoding. The sequence prediction model appears to be sensitive to certain features such as distance traveled. For example, changing the object from the "bucky ball" to the "bomb ball" decreases the model's confidence that the object rolling (though, the probe still correctly assigns a majority of the probability to \texttt{roll}). However, in this perturbation, the bomb ball gets stuck on its `cap' (Figure \ref{fig:all_objs}) and only completes one rotation.}
    \label{fig:extra_counterfactuals}
\end{figure*}

We start with a base set of 10 sequences: 5 with a sliding object (\texttt{cardboardBox\_03}) and 5 
with a rolling object (\texttt{BuckyBall}). We then create 20 minimal-edit perturbations to create a final set of 200 sequences.
We perturb the following features one at a time: \{\texttt{mass, force velocity, starting x position, starting z position, shape, angular rotation}\}. For most features, we generate 4 perturbations. For example, the x and z positions are altered by \{-2m, -1m, +1m, +2m\} where `m' is the Unity meter. All objects start with 1.14 units of mass and similar to the starting position variable, is altered by $1.14+(i\times.1)$ where $i$ is in the set \{-2, -1, 1, 2\}. For the shape parameter, we only change the 3D model used to generate the base sequence. For the sliding videos, we use \texttt{plate02} and \texttt{book\_0001c}. For the rolling videos we use \texttt{BombBall} and \texttt{modified Soccer Ball}. Note that we modify the \texttt{Soccer Ball} model that is in the train set, but modify the mass (1.14) and size of the model so that it is technically an unseen object. We chose to do this because we wanted to use a more plain spherical object, which was not an option for the remaining test objects. Angular rotation either perturbs the sequence by freezing the rotation along all axes (in the case of objects that normally roll) or replacing the physics collider with a sphere (causing the object to roll -- in the case of objects that tend to slide instead of roll). Figure \ref{fig:extra_counterfactuals} shows additional perturbations and a sliding object example of the counterfactual analysis.

\begin{figure*}[ht]
    \centering
    \includegraphics[scale=0.75]{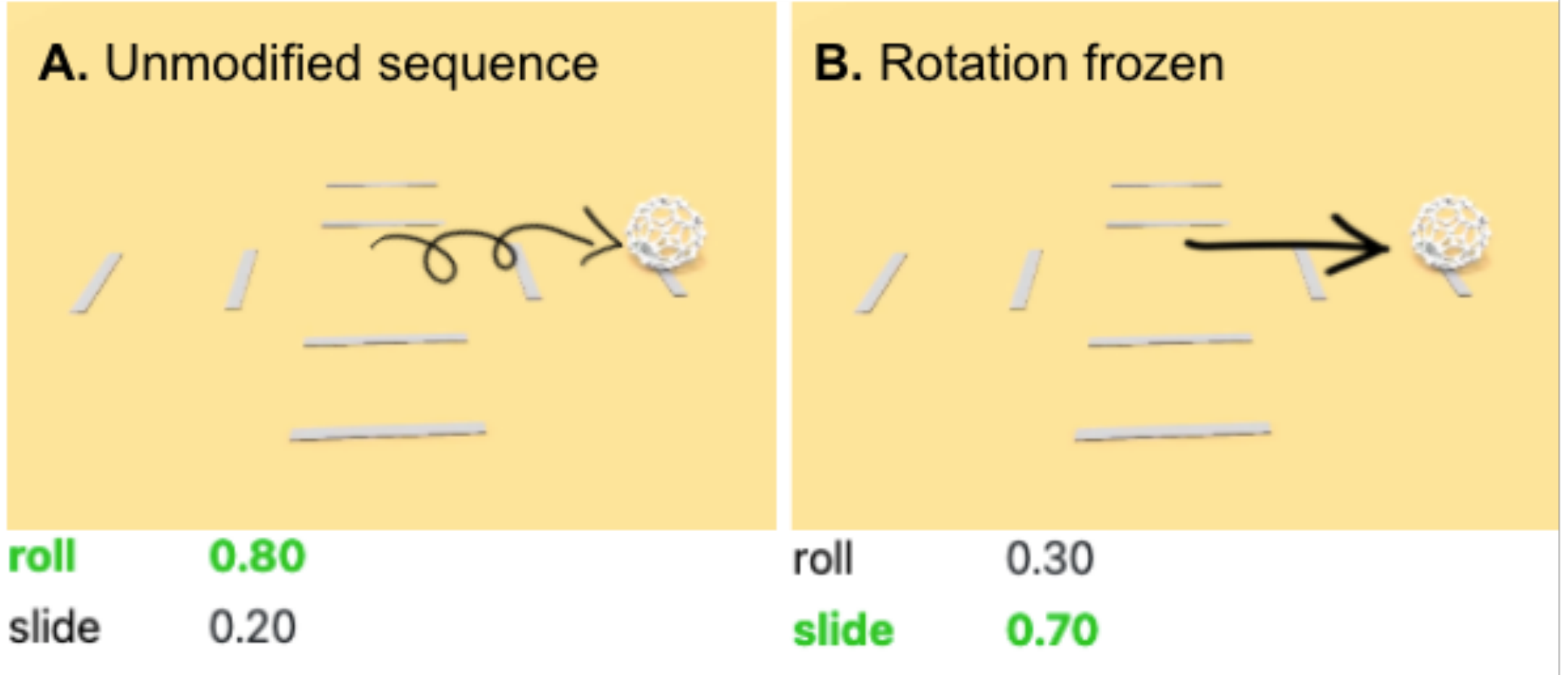}
    \caption{Left: a sequence generated with normal physics. Right: rotation locked, with all other physical properties of the interaction the same. Freezing the rotation such that the object slides causes the model to encode the action as a \texttt{slide} rather than a \texttt{roll}}
    \label{fig:demo_roll_ex}
\end{figure*}
\end{document}

%% file: sections/intro.tex
\section{Introduction}
Much of natural language semantics concerns events and their participants--i.e., verbs and the nouns with which they compose. Evidence from cognitive science \citep{borghi2009sentence, abstract_affs} and neuroscience \citep{sakreida2013abstract} suggests that grounding such words in perception is an essential part of linguistic processing, in particular suggesting that humans represent nouns in terms of their \textit{affordances} \citep{gibson1977theory}, i.e., the interactions which they support.
\begin{figure}[ht!]
    \centering
    \includegraphics[scale=.18]{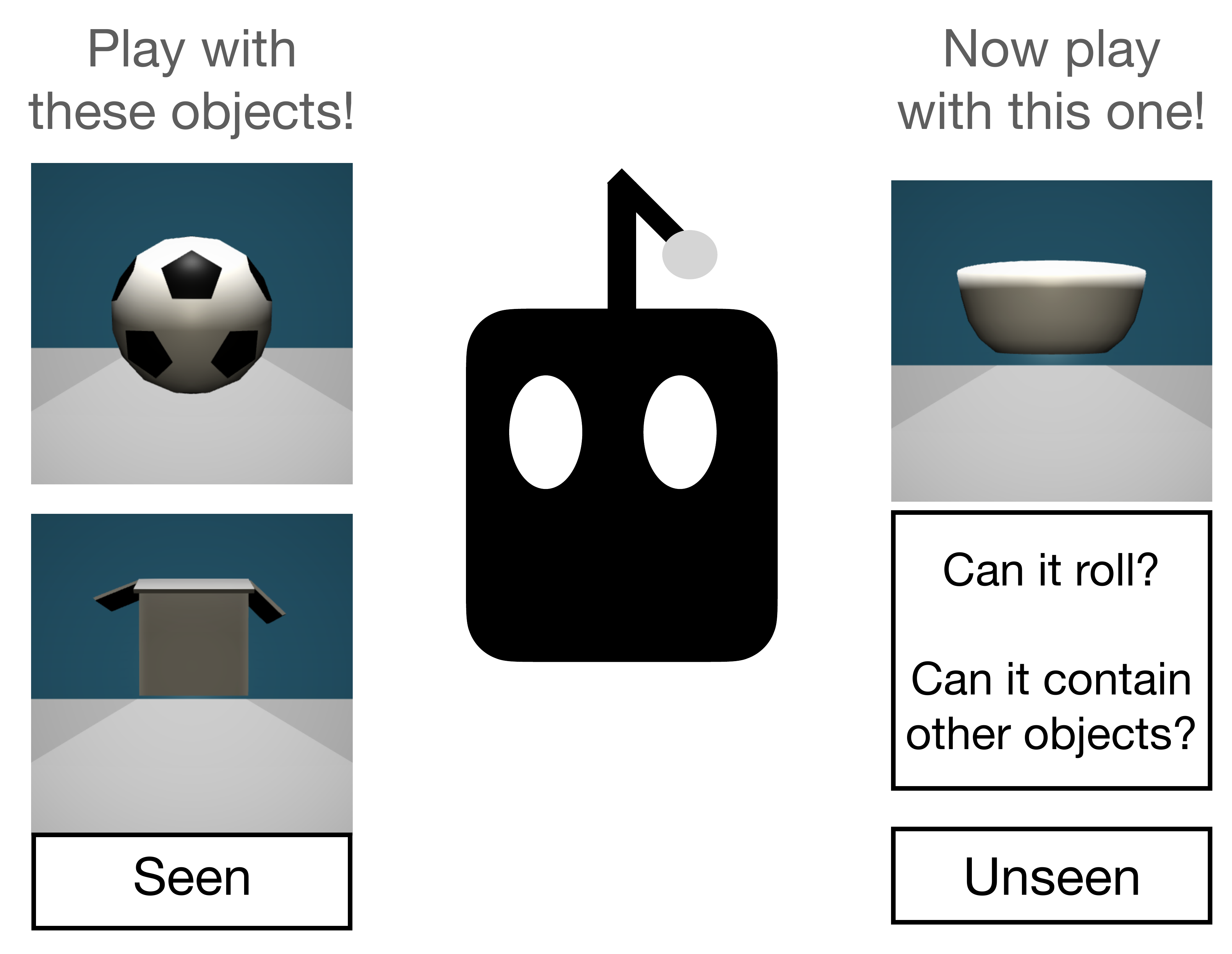}
    \caption{We investigate whether observing interactions with an object in a 3D environment encodes information about their affordances and whether this generalizes in the zero shot setting to unseen object types}
    \label{fig:motivating}
\end{figure}
Affordance-based representations have been argued to form the basis of formal accounts of compositional syntax and semantics \citep{steedman2002plans}. As such, prior work in formal semantics has sought to build grounded lexical semantic representations in terms of objects and their interactions in 3D space. For example, \citet{pustejovsky-krishnaswamy-2014-generating} and \citet{siskind2001grounding} represent verbs like \textit{roll} as a set of entailed positional and rotational changes specified in formal logic, and \citet{pustejovsky-krishnaswamy-2018-every} argue that nouns imply (latent) events--e.g., that \textit{cups} generally \textit{hold} things--which should be encoded as \textsc{telic} values within the noun's formal structure.
  
  \begin{figure*}[ht!]
    \centering
    \includegraphics[width=\linewidth]{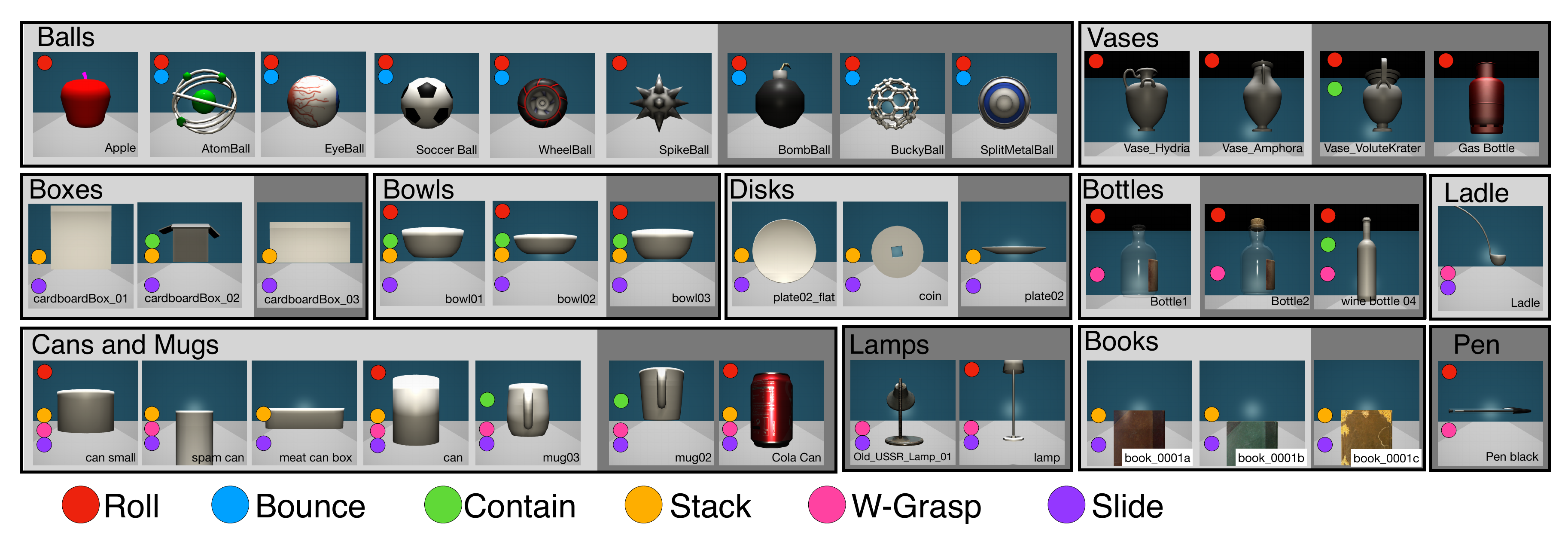}
    \caption{Objects uses for train (light gray) and test (dark gray). Colored dots indicate which affordances each object has.}
    \label{fig:seen_unseen}
\end{figure*}

Such work provides a compelling story of grounded semantics, but has not yet been connected to the types of large scale neural network models that currently dominate NLP. Thus, in this work, we ask whether such semantic representations emerge naturally as a consequence of self-supervised predictive modeling. Our motivation stems from the success of predictive language modeling at encoding syntactic structure. That is, if neural language models trained to predict text sequences learn to encode desirable grammatical structures \citep{kim2021testing,tenney2018you}, perhaps similar models trained to predict event sequences will learn to encode desirable semantic structures. To test this intuition, we investigate whether a transformer \citep{VaswaniAttention} trained to predict the future state of an object given perceptual information about its appearance and interactions will latently encode affordance information of the type thought to be central to lexical semantic representations. In sum:
\begin{itemize}
 \item We present a first proof-of-concept neural model that learns to encode the concept of affordance without any explicit supervision.
  \item We demonstrate empirically that 3D spatial representations (simulations) substantially outperform 2D pixel representations in learning the desired semantic features.
  \item We release the \ours\ dataset of 9.5K simulated object interactions and accompanying videos, and an additional 200K simulations without videos to support further research in this area.\footnote{\url{https://github.com/jmerullo/affordances}}
  \end{itemize}

\noindent Overall, our findings suggest a process by which grounded lexical representations--of the type discussed by \citet{pustejovsky-krishnaswamy-2014-generating} and \citet{siskind2001grounding}--could potentially arise organically. That is, grounded interactions and observations, without explicit language supervision, can give rise to the types of conceptual representations to which nouns and verbs are assumed to ground. We interpret this as corroborative of traditional feature-based lexical semantic analyses and as a promising mechanism of which modern ``foundation'' model \citep{foundation} approaches to language and concept learning can take advantage.

%% file: sections/dataset.tex
\section{Experimental Setup}
\label{sec:dataset}

\subsection{Objects and Affordances in \ours} \label{sec:affordances}
To collect a set of affordances to use in our study, we begin with lists of affordances and associated objects that have been compiled by previous work on affordance learning:
\citet{prost} provides on a small list of concrete actions for evaluating physical reasoning in large language models; \citet{rgbd_part_affs} provides a small list for training computer vision models to recognize which parts of objects afford certain actions; \citet{chao_semantic_affs} use crowdworkers\footnote{In the case of \citet{chao_semantic_affs}, we use a score $\ge4$ as positive label, as they do in their paper.} to judge noun-verb pairs and includes over 900 verbs that are both abstract and concrete in nature. We then filter this list down to only a subset of concrete actions that include objects which exist in the Unity asset store, since we use Unity simulated environments to build our training and evaluation data (\S\ref{sec:environment}). This results in a list of six affordances (\texttt{roll, slide, stack, contain, wrap-grasp,  bounce}) which are used to assign binary labels to each of 39 objects from 11 object categories (Figure \ref{fig:seen_unseen}; see also Appendix \ref{sec:appendix}).

\subsection{Representation Learning}
\label{sec:replearning}
We hypothesize that predictive modeling of object state will result in implicit representations of affordance and event concepts, similarly to how predictive language modeling results in implicit representations of syntactic and semantic types. Thus, for representation learning, we use a sequence-modeling task defined in the visual and/or spatial world. Specifically, given a sequence of frames depicting an object's trajectory, our models' objective is to predict the next several timesteps of the object's trajectory as closely as possible. We consider several variants of this objective, primarily differing in how the represent they frames (e.g., as 2D visual vs.\ 3D spatial features). These models are described in detail in Section \ref{sec:models}.

\subsection{Evaluation Task}

We are interested in evaluating which variants of the above representation learning objective result in readily-accessible representations of affordance and event concepts. To do this, we train probing classifiers \citep{analysis_nlp} on top of the latent representations that result from the representation learning phase. That is, we freeze the weights of our pretrained models and feed the intermediate representation for a given input from the encoder into a single linear-layer trained to classify whether the observed object has the affordance. We train a separate classifier probe for each affordance.

We construct train and test splits by holding out a fraction of the objects from each category. In some cases, the held-out objects are very similar to what has been seen in training (e.g., slightly different dimensions of boxes) and in other cases, the objects are visually very distinct (e.g., a wine bottle vs.\ a gas tank as instances of objects which afford both \texttt{roll} and \texttt{contain}). Figure \ref{fig:seen_unseen} shows our objects, affordances, and train-test splits.


\begin{figure*}[ht!]
    \centering
    \includegraphics[width=\linewidth]{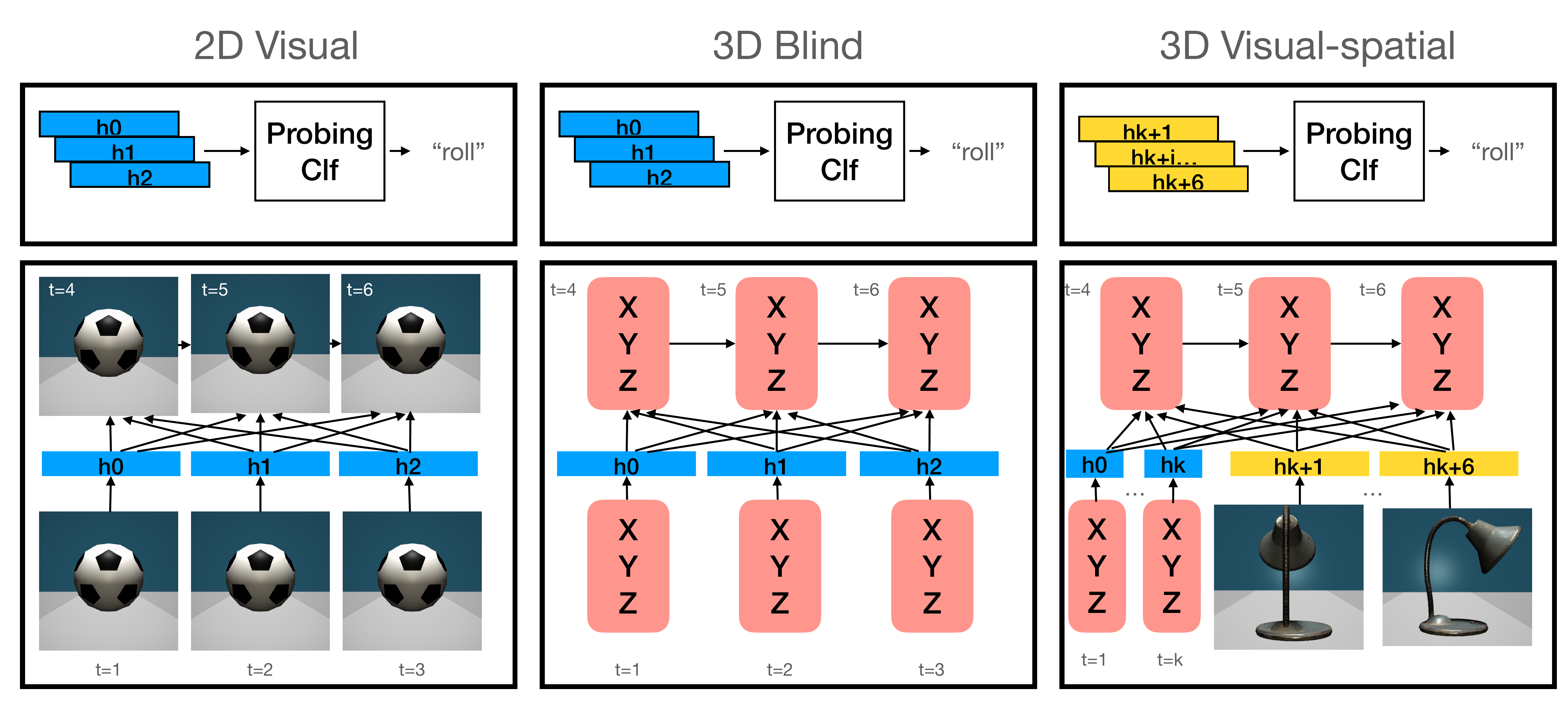}
    \caption{Model architectures. The model receives either images, 3D coordinates or both to make predictions. For the 3D models, the transformer encoder encodes the input sequence (and multiview images, if applicable) and the decoder predicts the rest of the sequence. For the 2D model (\texttt{RPIN}), a convolutional network extracts object-centric features ($h_i$) and interaction reasoning is performed over each to predict the next time steps.}
    \label{fig:full_model}
\end{figure*}

\subsection{\ours\ Environment and Data Collection}
\label{sec:environment}
The \ours\ dataset consists of simulations of interactions with a variety of 3D objects in the Unity game engine\footnote{\url{https://unity.com/}}.
Our data is collected in a flat empty room using the Unity physics engine on the above-described 39 objects. For each sequence, an object is instantiated at rest on the ground. A random impulse force--either a `push' flat along the ground, or a `throw' into the air--is exerted on the object. We only exert a single impulse on an object per sequence. The sequence ends when the object stops moving or after 4 seconds elapse.

We record the coordinates of the object in 3D space at a rate of 60 frames per second. Specifically, each sequence is defined by the coordinates describing the object's 3D position in space $P=\{p_1, ... , p_t\}$ for $t$ timesteps. Since we care about capturing the manner in which the object travels and rotates through space, $p_i$ contains 9 distinct 3D points around the object: 8 corners around an imaginary bounding box and the center point of that bounding box (see Appendix \ref{sec:appendix} for a visual aid). Simultaneously we collect videos of each interaction from a camera looking down at a 60 degree angle towards the object that we will use to train our 2D vision based model. Each image in the videos is collected at a resolution of 384x216 pixels. We filter videos where the object leaves the frame. Overall, this process results in 2,376 training sequences
and 9,283 evaluation sequences. Due to computational constraints, we decided to focus on collecting as many evaluation examples as possible to make comparison to spatial models easier and more accurate. It may be the case that adding more data creates stronger representations, but even with this smaller training set, we see high test time performance on the visual dynamics task.
All our data are publicly available at \url{https://github.com/jmerullo/affordances}.

\subsection{Assumptions and Limitations}
This work serves as initial investigation of our hypothesis about representation learning for affordances (\S\ref{sec:replearning}). We use simple simulations which involve only a single object. Thus, we expect that our setup makes some affordances (\texttt{roll, slide, bounce}) more readily available than others (\texttt{contain, stack, w-grasp}). For example, our models likely will observe objects rolling during pretraining, but will never observe objects being stacked on top of one another. 
However, during evaluation, we will assess how well the model's internal representations encode both types of affordances. This is intended. Our hypothesis is that, to a large extent, these affordances are a function of the relationship between the shape\footnote{In fact, \citet{gibson1977theory}'s original theory of affordances defined them to be purely-perceptual, without even depending on internal processing and representation. We do not endorse this view in general; we are enthusiastic about future work which involves richer internal processing (e.g., interaction and planning) during pretraining. See \citet{toafford} for a review of the various definitions and interpretations of the term that have been used in different fields and \citet{nonperceptualaffs} for an argument that affordances are not solely perceptual. That said, this basic-perception approach is helpful starting point for understanding the relationship between pretraining and affordance representations. } of the objects and the physics of how those objects behave in our simulation. For example, we expect that long, thin \texttt{grasp}-able objects will display different trajectories than will wide, round objects that cannot be grasped. Thus, we expect that a model trained to predict object trajectories can encode differences that map onto affordances such as \texttt{grasp} or \texttt{contain}, even without observing those actions \textit{per se}. Given initial promising results (\S\ref{sec:experiments}), we are excited about future work which extends the simulation to include richer multi-object and agent-object interactions, which likely would enable learning of more complex semantic concepts.

%% file: sections/model.tex
\section{Models}\label{sec:models}
We consider two primary variants of the representation learning task described in Section \ref{sec:replearning} which differ in how they represent the world state--i.e., using 2D visual data (\S\ref{sec:rpin_description}) vs.\ using 3D visual-spatial data (\S\ref{sec:multiview_images}). To provide additional insights into performance differences, we also consider two ablations in the 3D model (\S\ref{sec:notraining}), one that removes visual information and one that further removes pretraining altogether. These models are summarized in Figure \ref{fig:full_model}.

\subsection{2D Visual Model}
\label{sec:rpin_description}
We first consider a standard computer-vision (CV) approach for our defined representation learning objective. For this, we use a Region Proposal Interaction Network (\texttt{RPIN}) proposed in \citet{rpin}. We choose to use \texttt{RPIN} because it was designed to solve a task very similar to ours--i.e., object tracking over time--and has access to object representations via bounding boxes provided as supervision during training. Using a model with access to explicit object representations ensures that we are not unfairly handicapping the CV approach (by requiring it to learn the concept of objects from scratch) but rather are analyzing the relative benefits of a 2D CV approach vs.\ a 3D spatial data approach for latently encoding semantic event and affordance concepts.

We train the model with similar settings to those \citet{rpin} used to train on the Simulated Billiards dataset, but with some small differences. For example, we subsample our frames to be coarser-grained to encourage learning of longer-range dependencies. 
Exact details and explanations of other parameter differences can be found in Appendix \ref{sec:rpin_appendix}.

To probe object representations for affordance properties, we take the average of the hidden representations--i.e., the model's representations just prior to predicting explicit bounding box coordinates on the screen.

\subsection{3D Visual-spatial Model} \label{sec:multiview_images}

\subsubsection{Full Model}

Recent work has argued that models based directly on 3D game engine data are more cognitively plausible for modeling verb semantics \citep{dylan-starsem}.
In this spirit, we consider a model that learns to encode the objects visual appearance jointly with predicting the objects' behavior in 3D space. Specifically, our model is trained with both an object loss and a trajectory loss as follows.

To model the 3D trajectory, the model encodes a sequence $P$ containing positions $\{p_1, p_2, ..., p_t\}$. As described in Section \ref{sec:dataset}, each position $p_i$ contains 9 distinct points corresponding to the object center and the 8 corners of the rectangular bounding box encapsulating the object. We use a single linear layer to project the 27D (9 3D points) input coordinate vectors to the embedding dimension of a transformer \citep{VaswaniAttention}. The transformer is then fed the first $t-k$ timesteps where $k\geq1$. We treat $k$ as a hyperparameter, and find that a value of $k=8$ or $k=16$ tends to work the best.
Our model is trained to minimize the Mean Squared Error (MSE) computed against the true future location of the object, summed over all of the predicted points. 

To model the object appearance, we give the model access to a static view of the object at rest. 
We use ResNet-34 \citep{resnet} to encode the object's \textit{multiview}--i.e., images of the object's six faces, one from each side of the object--denoted as $I$, and pass these as additional inputs to the model, separated by a \textsc{SEP} token. The transformer encoder encodes the sequence $P$ and $I$ together, and the transformer decoder predicts the object's next several positions in space. 
To encourage the model to connect the sequence and image representations, we randomly (50\% of the time) replace the object in $I$ with an object with different affordances and add an auxiliary loss in which the model must classify whether the object was perturbed. We add a linear binary classification layer on top of a CLS token to perform this task, and add the cross entropy loss of this objective to our MSE loss for the trajectory objective.


The hidden representation we use for probing experiments is the average pooled transformer encoder output of the multiview tokens only.

\begin{figure*}[ht!]
    \centering
    \includegraphics[width=\linewidth]{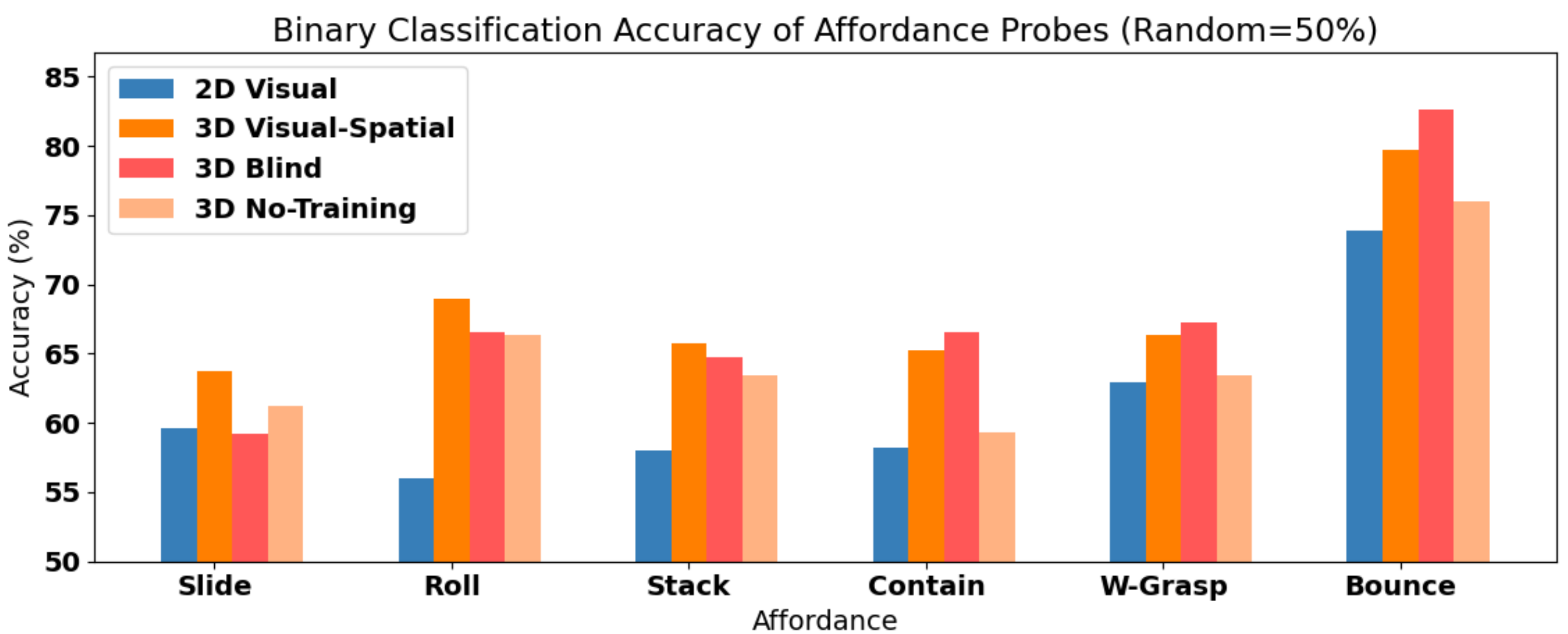}
    \caption{Results for predicting affordances of objects given frozen hidden states of 2D and 3D sequence prediction models. Test sets are balance so that random guess achieves 50\%. 3D models (even ablated variants) outperform the 2D computer vision models across the board.}
    \label{fig:probe_results}
\end{figure*}
\subsubsection{Ablation Models}

To better understand which aspects of the above model matter most, we also train and evaluate two ablated variants.  

\paragraph{Without Visual Information (3D Blind).} \label{sec:base_model}

Our 3D Blind model is like the above, but contains no multiview tokens or associated loss. That is, the model is trained only on the 3D positional data, using an MSE loss to predict the future location of the object. For probing, we average the transformer encoder outputs across all timesteps and feed the single averaged emebedding into the probing classifier. This model provides insight into how well the physical behavior alone, with \textit{no visual inputs}, encodes key features for determining affordances, such as shape and material.

\paragraph{Without Pretraining (No-Training).}
\label{sec:notraining}
Gibson believed that understanding affordances only required raw perception, without need for mental processing. Given how saliently actions like rolling and sliding are encoded in 3D coordinates (Figure \ref{fig:corner_roll_slide}), it is reasonable to ask how much benefit our pretraining objective provides for encoding affordance information. To test this, we evaluate a model that is identical to the 3D Blind model, but contains only randomly initialized encoder weights (i.e., which are never set via pretraining). If the pretraining task encodes affordance structure the way we hypothesize, the randomly initialized model should perform much worse than the trained 3D Blind variant. We refer to this model simply as the No-Training model.

%% file: sections/experiments.tex
\begin{figure}[ht!]
    \centering
    \subfloat[\centering
    3D Blind model
    ] {{\includegraphics[height=2.7cm, width=3.7cm]{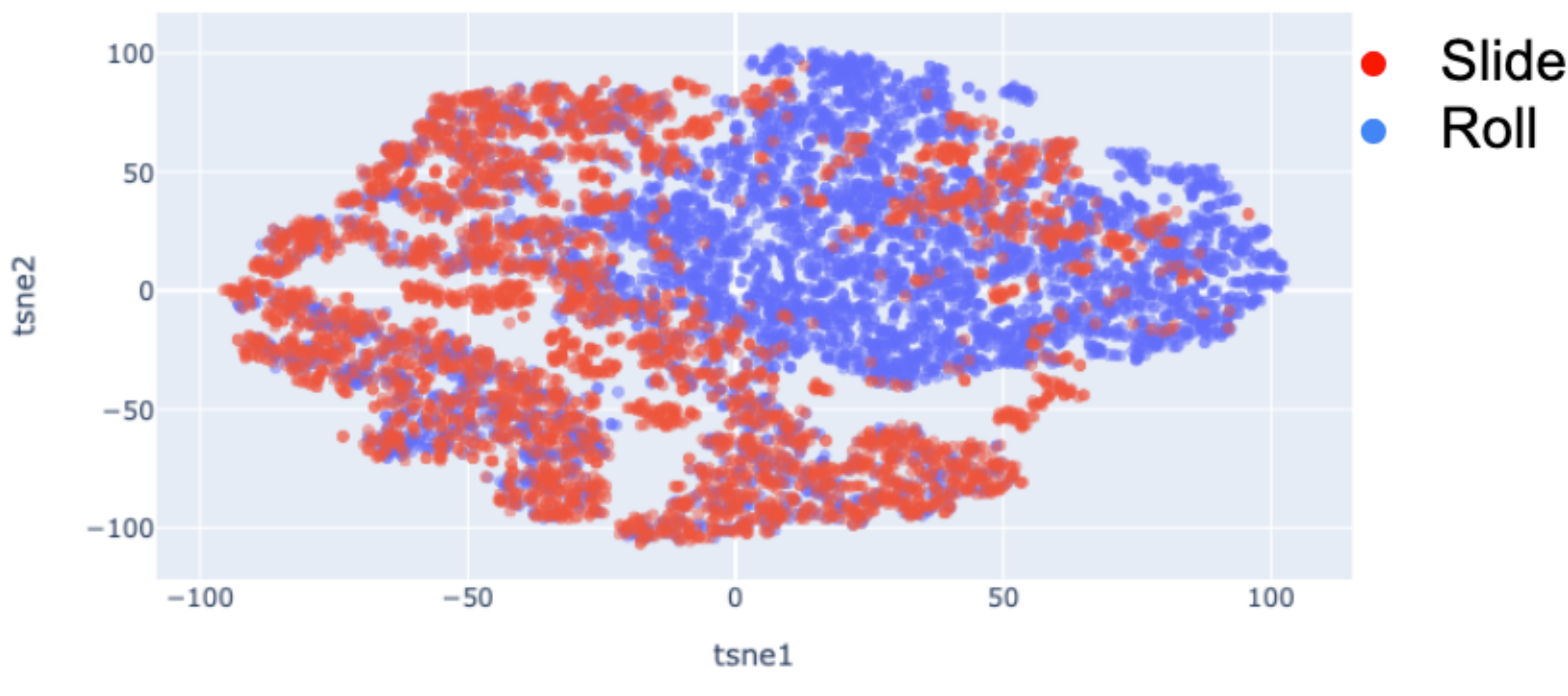} }}%
    \subfloat[\centering 
    3D Visual-spatial model
    ]
    {{\includegraphics[height=2.7cm, width=3.7cm]{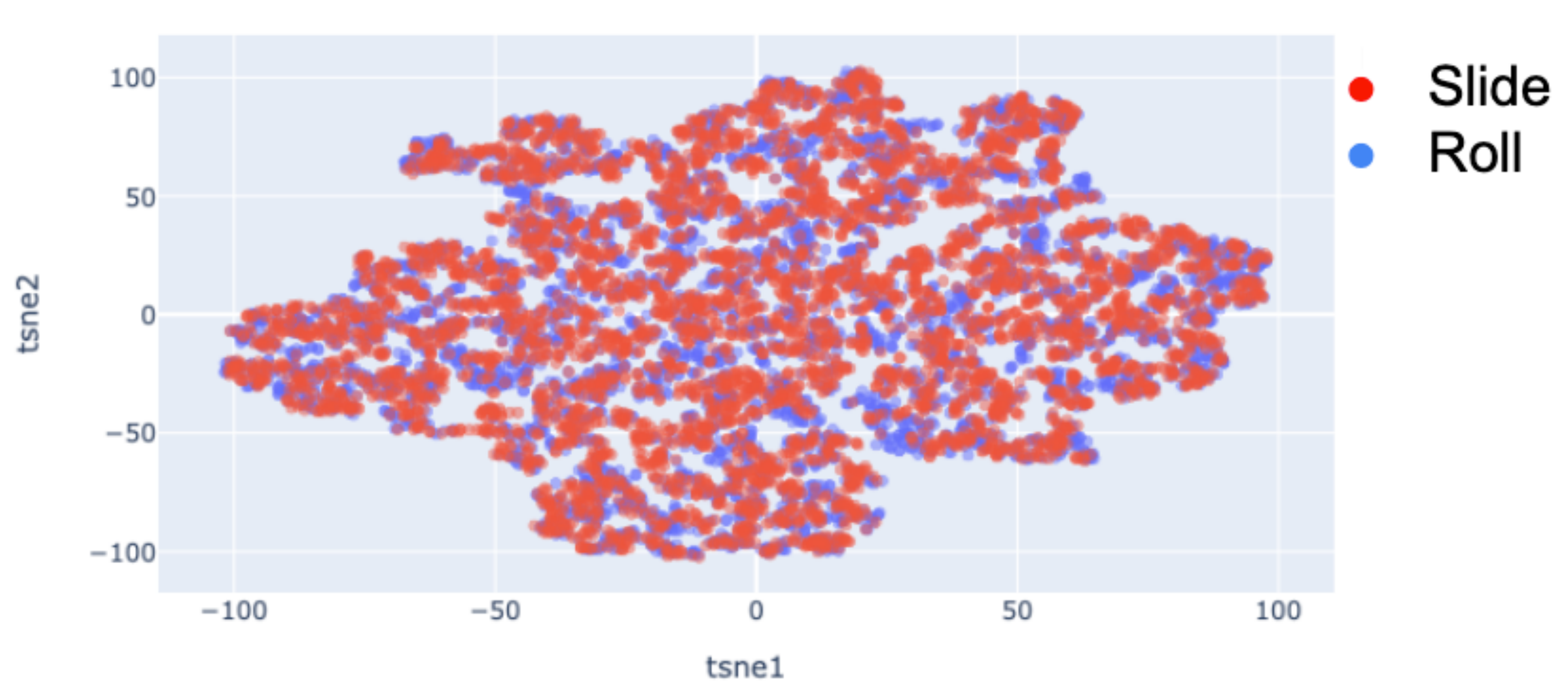} }}%
    \qquad
    \subfloat[\centering
    3D No-Training
    ] {{\includegraphics[height=2.7cm, width=3.7cm]{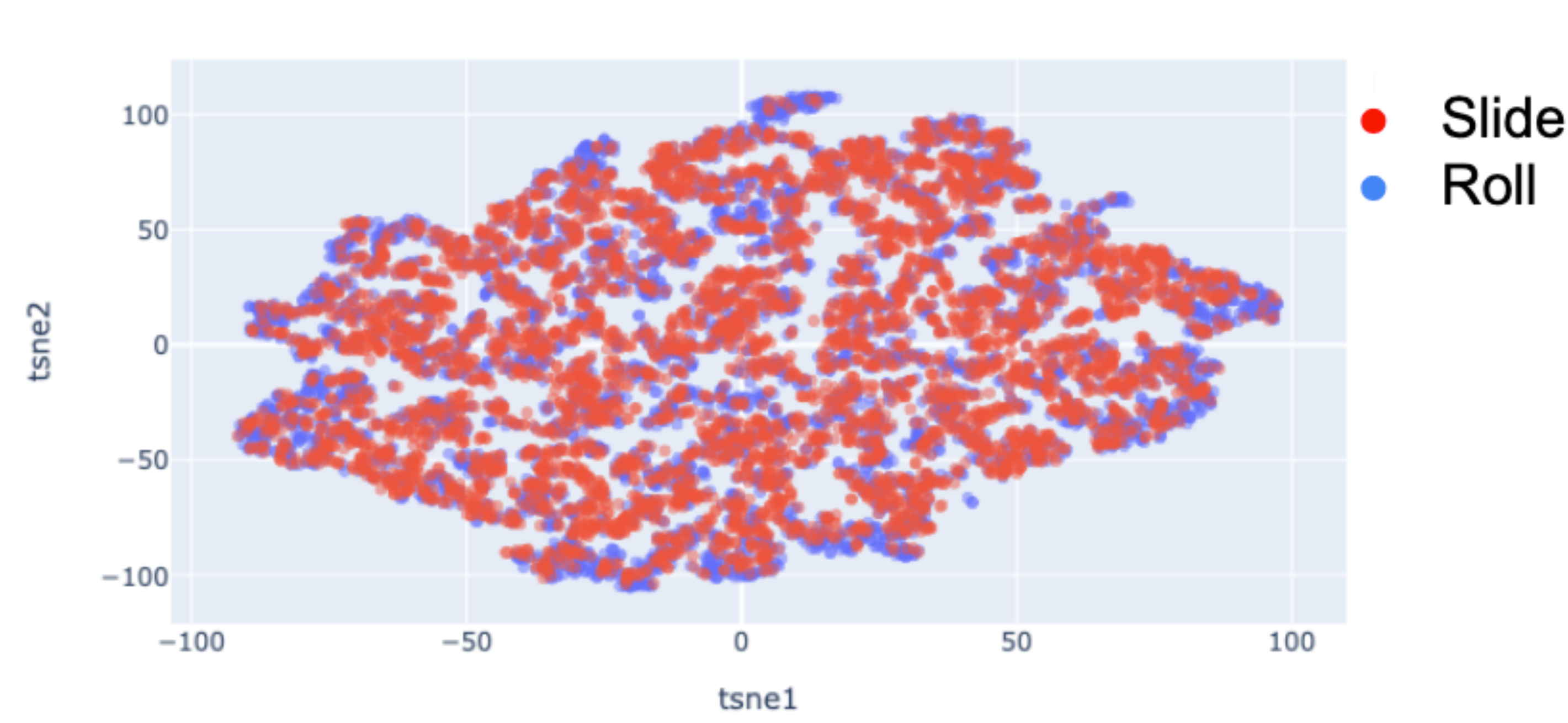} }}%
    \subfloat[\centering 
    2D Computer-Vision model (RPIN).
    ]
    {{\includegraphics[height=2.7cm, width=3.7cm]{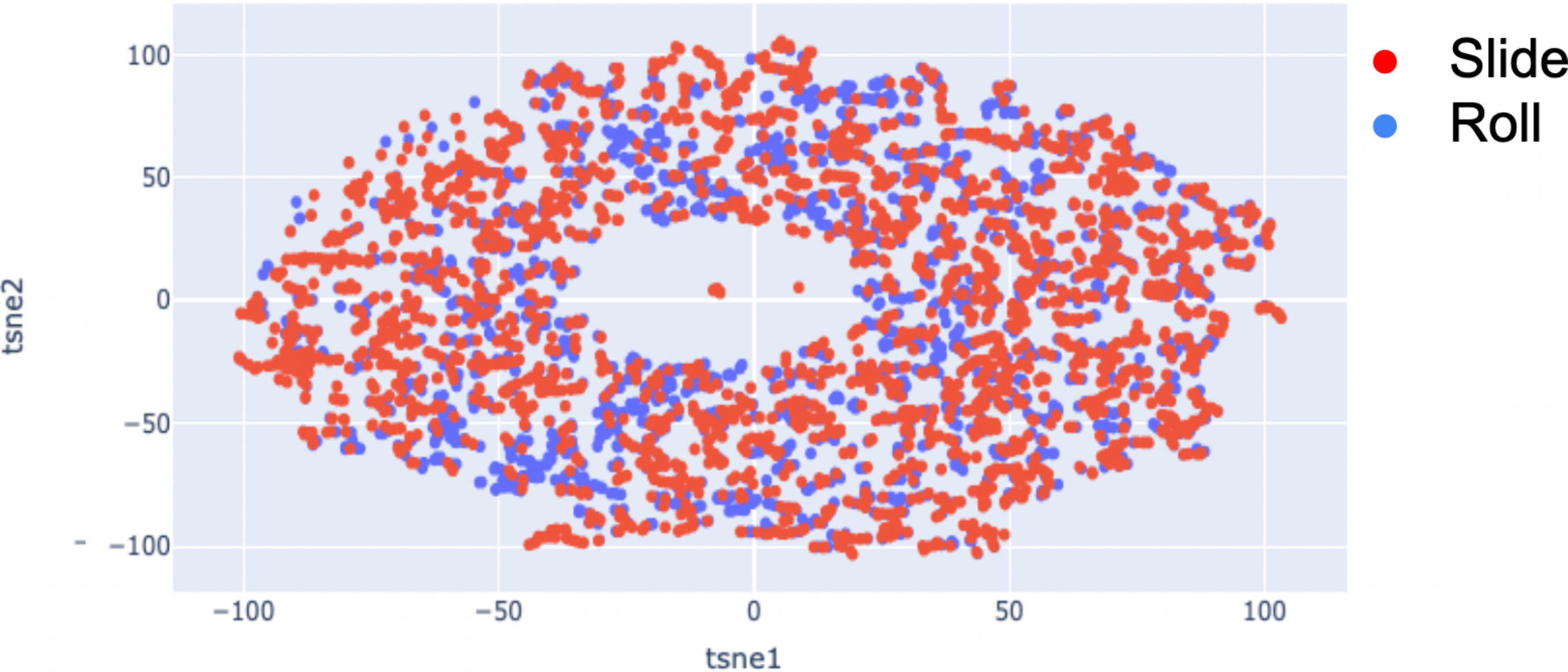} }}%
    \caption{t-SNE projections of model representations of sliding (red) vs.\ rolling (blue) objects.}
    \label{fig:unity_vs_rpin_clusters}%
\end{figure}

\section{Results}\label{sec:experiments}
Figure \ref{fig:probe_results} shows our primary results. Overall, the 3D Visual-spatial model substantially outperforms the 2D Vision-only model across all affordances, often by a large margin (4--11 percentage points). We also see, perhaps unexpectedly, that the 3D pretrained representations encode information about affordances even when the associated actions are not explicitly observed. For example, the model differentiates objects that can \texttt{stack} and objects that can \texttt{contain} other objects from those that cannot, even though the model has not directly observed objects being stacked or serving as containers during training. This result points to the richness of the physical information that is required to perform the pretraining task of next-state prediction.\footnote{We note that, unintuitively, \texttt{stack} and \texttt{contain} probes generally outperform \texttt{slide} probes. One reason may be because our data are labeled by object rather than by individual interaction. For example, although an object \textit{typically} slides, it's not hard to imagine scenarios where a cardboard box might roll over. This is not the case for affordances like \texttt{stack} and \texttt{contain}. In the rolling cardboard box example, the sharp edges of the box and the distinct way it rolls is still indicative of the object being stackable.}

Looking more closely at the ablated variants of the 3D model, we see that most of the gains are from the 3D input representation itself. That is, the 3D No-Training model--which does not include visual information and does not even include pretraining--outperforms the CV baseline in all cases, and often substantially. Pretraining on top of the 3D inputs often (but not always) yields performance gains. Pretraining with visual information does not provide a clear benefit over pretraining on the spatial data alone--i.e., visual information leads to performance gains on three affordances (\texttt{slide}, \texttt{roll}, and \texttt{stack}) and losses on the other three (\texttt{contain}, \texttt{w-grasp}, and \texttt{bounce}).



\begin{figure}[ht!]
    \centering
    \includegraphics[scale=.45]{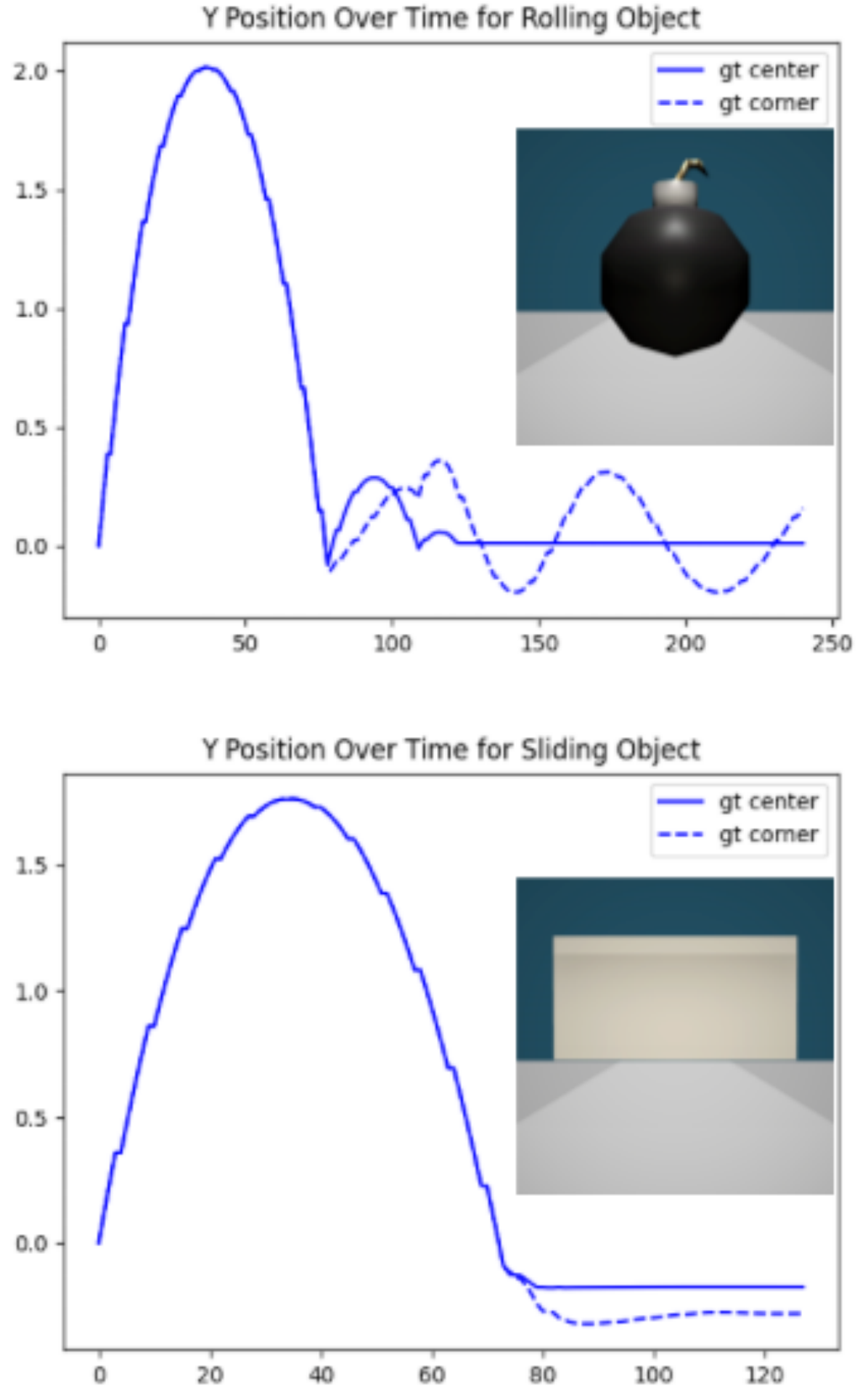}
    \caption{Visualization showing how 3D coordinate data clearly distinguishes a rolling object from a sliding one, making it easier for a model to learn the difference between the two.}
    \label{fig:corner_roll_slide}
\end{figure}

\section{Qualitative Analysis}

In order to better understand the nature of the affordance-learning problem, we run a series of qualitative analyses on the trained models. We focus our analysis on the pair of affordances \texttt{roll} vs.\ \texttt{slide}. These are verbs have received significant attention in prior literature \citep{pustejovsky-krishnaswamy-2014-generating,levin1993english} since they exemplify the types of manner distinctions that we would like lexical semantic representations to encode. 


We first compare the 2D video vs.\ 3D simulation variants of our pretraining objective. Figure \ref{fig:unity_vs_rpin_clusters} shows a t-SNE projection of the sequence representations from all four models, labeled based on if the object affords rolling or sliding. We find that object representations from the 3D Blind model cluster strongly according to the distinction between these two concepts. The trend is notably not apparent in the No-Training model. Figure \ref{fig:corner_roll_slide} demonstrates why spatial data pretraining may encourage this split. In the example shown, we take two thrown objects from our dataset--one round and one not round--and track the height of the center point of the object and one of the corners of the object bounding box. When they hit the ground the center point stays relatively constant as it moves across the floor in both, but in the rolling action, the corner point moves up and down as it rotates around the center point. Since this is so distinguishable given the input representation, the model is better able to differentiate these concepts.

It may be that the next state prediction task facilitates learning the slide vs.\ roll distinction in the 3D Blind setting. However, the same pattern is not present in the 3D Visual-spatial model (which also predicts next state). One possibility is that the presence of visual information competes with the 3D information, and as a result the joint space does not encode this distinction as well as the 3D space alone. Designing more sophisticated models that incorporate visual and spatial information and preserve the desirable features of both is an interesting area for future work. 


\subsection{Counterfactual Perturbations}
\label{sec:error_analysis}

An important aspect of lexical semantics is determining the \textit{entailments} of a word--e.g., what about an observation allows it to be described truthfully as \texttt{roll}? Thus, in asking whether affordances are learned from next-state-prediction pretraining, it is important to understand not just whether the model can differentiate the concepts, but whether it differentiates them for the ``right'' reasons.

We investigate this using counterfactual perturbations of the inputs as a way of doing feature attribution, similar in spirit to prior work in NLP \citep{huang-etal-2020-reducing} and CV \citep{goyal2019counterfactual}. Specifically, we create a controlled dataset in which, for each of 10 interactions, we generate 20 minimal-pairs which differ from their originals by a single parameter of the underlying physics simulation. The parameters we perturb are \{\texttt{mass, force velocity, starting x position, starting z position, shape, angular rotation}\}. For example, given an instance of a lamp rolling across the floor, we would generate one minimally-paired example in which we only change the mass of the lamp, and another the same as the original except it does not exhibit any angular rotation, and so forth for each of the parameters of interest. More implementation details are given in Appendix \ref{sec:demo_appendix}.

\begin{figure}[ht!]
    \centering
    \includegraphics[width=.45\textwidth]{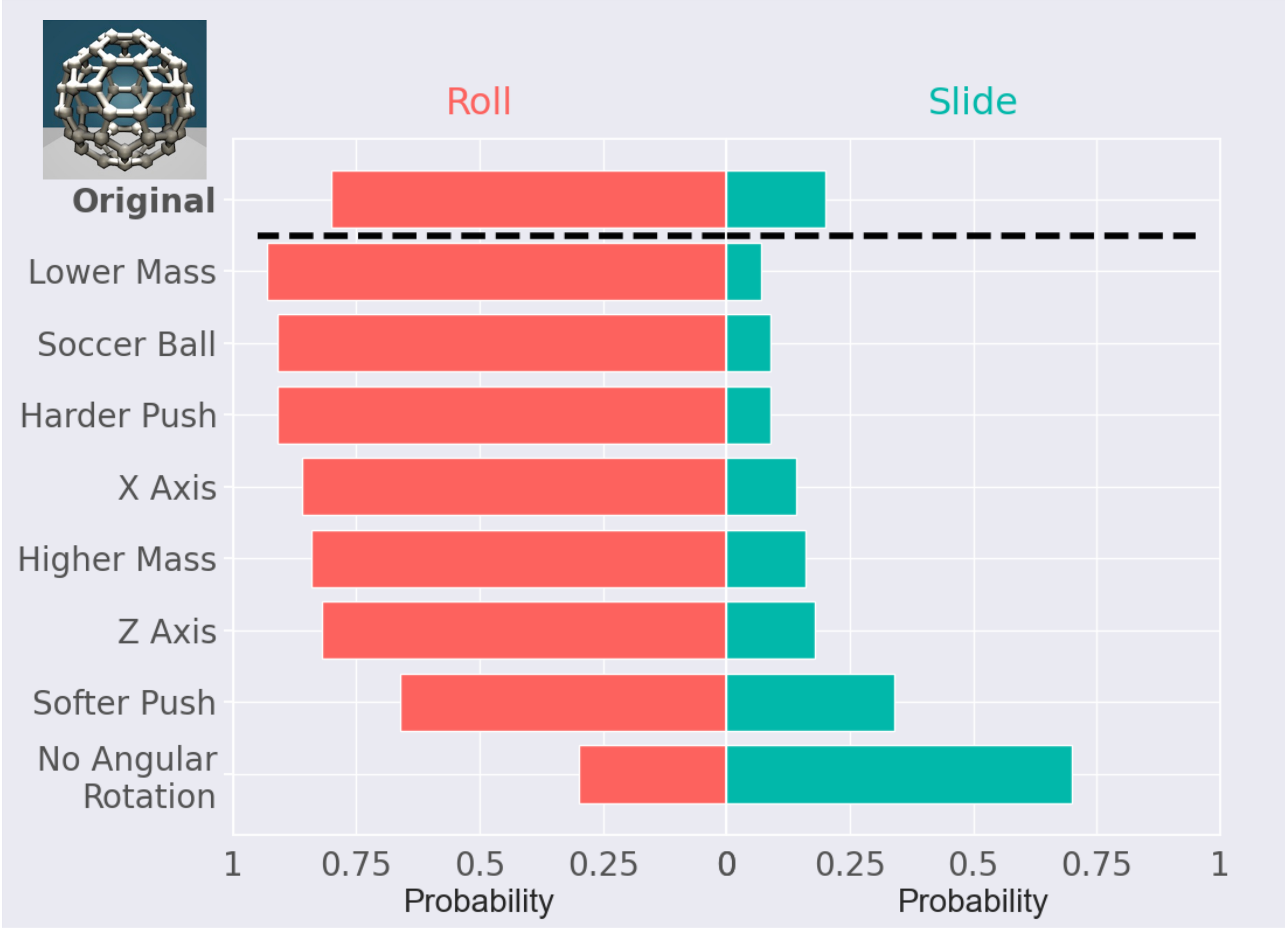}
    \caption{Change in predicted probability of the encoding of a round object affording \texttt{slide} after generating the interaction again with one feature changed (See Appendix \ref{sec:demo_appendix} for a visualization)}
    \label{fig:demo_roll_probs}
\end{figure}

We use our pretrained \texttt{slide} probe to classify the representations from each sequence as either rolling or sliding, and compare the effect of each perturbation on the model's belief about the affordance label. Figure \ref{fig:demo_roll_probs} shows the resulting belief changes for several of the perturbed parameters. We see that changing the angular rotation of an otherwise identical sequence has the greatest effect on whether an instance is deemed to afford \texttt{roll}ing. This is an encouraging result, as it aligns well with standard lexical semantic analyses: i.e., generally, \texttt{roll} is assumed to entail rotation in the direction of the translation. 

However, our analysis also reveals that the models rely on some spurious features which, ideally, would not be part of the lexical semantic representation. For example, the 3D blind model is affected by the travel distance the object. If we increased the mass or decreased the force applied to a rolling object, such that it only moved a small distance or rotated a small number of times, the model was less inclined to label the instance as rolling; though this was usually not by enough to have the undesirable effect of flipping the prediction.
Intuitively this makes sense given the model's training data: rolling objects tend to travel a greater distance than sliding objects. An interesting direction for future work is to investigate how changes in pretraining or data distribution influence which features are encoded as ``entailments'', i.e., key distinguishing features of a concept's representation. 

%% file: sections/related.tex
\section{Related Work}\label{sec:related}

\subsection{Lexical Semantics and Cognitive Science}\label{sec:related:affordances}

In formal semantics, there has been significant support for the idea that motor trajectories and affordances should form the basis of lexical semantic representations \citep{pustejovsky-krishnaswamy-2014-generating,siskind2001grounding,steedman2002plans}. Such work builds on the idea in cognitive science that simulation lies at the heart of cognitive and linguistic processing \citep{feldman2008molecule,bergen2007spatial,bergen2012louder}.
For example, \citet{borghi2009sentence} argue that language comprehension involves mental simulation resulting in a "motor prototype" which encodes stable affordances and affects processing speed for identifying objects. 
\citet{teliccomponent} point to such simulation as a factor in determining surprisal of affordances depending on linguistic context. Similar arguments have been made based on evidence from fMRI data \citep{sakreida2013abstract} as well as processing in patients with brain lesions \citep{strokeactionverbs}. 
It is worth noting that there is debate on the general nature of affordances in humans. For instance, \citet{nonperceptualaffs} argues that affordances are not solely perceptual. We view our work as being compatible with this more general view of affordances, in which direct perception plays a role, but not the only role, in concept formation.

\subsection{Affordances in Language Technology}

The idea of affordances has been incorporated into work on robotics \citep{toafford,affrobosurvey}. \citet{kalkan2014verb,ugur2009predicting} build a model of affordances based on (object, action, effect) tuples, but focus only on start and end state, and do not encode anything about manner. Relatedly, \citet{nguyen2020robot} connects images of objects to language queries describing their uses.

Affordances are also well studied for text understanding tasks. \citet{coercion} discuss the importance of affordances in disambiguating meaning of sentences such as ``we finished the wine".  Other neural net based approaches for affordance learning have relied on curated datasets with explicit affordance labels for each object \citep{chao_semantic_affs, affordancenet}. Sometimes, affordance datasets leverage multimodal settings such as images \citep{rgbd_part_affs}, or 3D models and environments \citep{EmbodiedBert,visual_grasping,thoraffs}, but require annotations for every object. In contrast, our model learns affordances in an unsupervised manner, and unlike \citet{fulda2017rock}, \citet{affordance_srl}, \citet{mcgregor-lim-2018-affordances}, and \citet{aff_unsupervised} which extract affordance structure automatically from word embeddings alone, our model learns from interacting with objects in a 3D space, grounding its representations to cause-and-effect pairs of physical forces and object motion.

\subsection{Physical Commonsense Reasoning}
There has been success in building deep learning networks that reason about object physics by learning to predict their trajectories. These can be broken up into either predicting points in 3D space given object locations (like our approach, e.g. \citet{Mrowca2018FlexibleNR}, \citet{se3-nets_2017}, \citet{InteractionNFBattaglia2016}, \citet{billiards_physics}, \citet{ye2018interpretable}, \citet{rempe2020predicting}) or inferring future bounding box locations of objects in videos \citep{video_object_tracking,affordancenet,rpin,ding2021attention}. Both approaches have been successful in encoding complex visual and physical features of objects. We focus on training with 3D simulations, but also test a visual dynamics model \citep{rpin} to compare the affordance information that is encoded from spatial vs. visual data.

More broadly, we contribute to a line of work on building non-linguistic representations of lexical concepts \citep{benchmarking_csr}.
Explicit attempts at grounding to the physical world ground language in 2D images or videos (i.e., pixels) \citep{action2vec,shapestacks}, despite the fact that recent work suggests that text and video pretraining offers no boost to lexical semantic understanding \citep{yun2021does}. Such efforts motivate the creation of large datasets such as \citet{visualgenome}, \citet{SituationRec}, and \citet{visualsrl}, which require in-depth human provided annotations that provide a limited list of semantic roles of objects. 

Our approach is most directly related to prior work that learns in interactive, 3D settings \citep{ispy,dylan-starsem}. Especially related are \citet{thoraffs} and  \citet{zellers_piglet_2021}. However, their models do not directly ground to the physical phenomena (e.g., entailed positional changes). Instead, they use a symbolic vocabulary of object state changes, whereas our model learns from unlabeled interactions.

%% file: sections/conclusion.tex
\section{Conclusion}

We propose an unsupervised pretraining method for learning representations of object affordances from observations of interactions in a 3D environment. We show that 3D trajectory data is a strong signal for grounding such concepts and performs better than a standard computer vision approach for learning the desired concepts. Moreover, we show through counterfactual analyses that the learned representations can encode the desired entailments--e.g., that \texttt{roll} entails axial rotation.

Our work contributes to an existing line of work that seeks to develop lexical semantic representations of nouns and verbs that are grounded in physical simulations. We advance this agenda by offering a way in which modern ``foundation model'' approaches to visual and linguistic processing can in fact be corroborative of traditional feature-based approaches to formal lexical semantics. Our results suggest a promising direction for future work, in which pretraining objectives can be augmented to include richer notions of embodiment (e.g., planning, agent-agent interaction) and consequently encoder richer lexical semantic structure (e.g., presuppositions, transitivity).  